# Proximal Symmetric Non-negative Latent Factor Analysis: A Novel Approach to Highly-Accurate Representation of Undirected Weighted Networks

Yurong Zhong, Zhe Xie, Weiling Li, and Xin Luo

**Abstract**—An Undirected Weighted Network (UWN) is commonly found in big data-related applications. Note that such a network's information connected with its nodes, and edges can be expressed as a Symmetric, High-Dimensional and Incomplete (SHDI) matrix. However, existing models fail in either modeling its intrinsic symmetry or low-data density, resulting in low model scalability or representation learning ability. For addressing this issue, a Proximal Symmetric Nonnegative Latent-factor-analysis (PSNL) model is proposed. It incorporates a proximal term into symmetry-aware and data density-oriented objective function for high representation accuracy. Then an adaptive Alternating Direction Method of Multipliers (ADMM)-based learning scheme is implemented through a Tree-structured of Parzen Estimators (TPE) method for high computational efficiency. Empirical studies on four UWNs demonstrate that PSNL achieves higher accuracy gain than state-of-the-art models, as well as highly competitive computational efficiency.

——————————— ◆ ———————————

## I. INTRODUCTION

IN big data-related applications such as predicting interactions among proteins [10, 11], an Undirected Weighted Network (UWN) is a common occurrence. This network comprises numerous nodes and observed edges, which can be represented as a Symmetric, High-Dimensional and Incomplete (SHDI) matrix. Evidently, an SHDI matrix possesses the following characteristics:
1. It is symmetric;
2. Its entities set is large;
3. Its most data entries are missing; and
4. Its data are commonly nonnegative like recommender system's ratings [23, 29-31].

Despite its incompleteness, an SHDI matrix contains valuable knowledge such as potential communities in a social network [7-9, 33]. Therefore, designing an analysis model that accounts for SHDI's characteristics and extracts such hidden knowledge is a crucial issue.

For effectively dealing with analysis of the target SHDI matrix generated by an UWN, recent studies' proposed models can mainly be split into three general categories: 1) Neural Network (NN) models [1, 2, 3]. For example, Sedhain *et al.*'s AutoRec [1] and He *et al.*'s LightGCN [2] are able to extract nonlinear latent factors (LFs) from the target matrix. In spite of attaining nonlinear LFs, they do not consider symmetry and incompleteness of an SHDI matrix; 2) Symmetric Nonnegative Matrix Factorization (SNMF) models [4, 5, 12]. For example, He *et al.*'s *β*-SNMF [4] and Yang *et al.*'s GSNMF [5] adopt one unique nonnegative latent factor (LF) matrix to precisely describe the symmetry and nonnegativity of an SHDI matrix. However, they also do not consider incompleteness of an SHDI matrix, i.e., they need to prefill its missing data before their training, thereby resulting in unnecessary high storage and computational cost; 3) Nonnegative Latent Factor (NLF) models [6, 48]. For example, based on a standard NLF model's data density-oriented objective function that is built by known data of an incomplete matrix [6], Luo *et al.*'s NIR [48] incorporates imbalanced information of the target matrix into NLF for better generalization. Note that they can efficiently handle an SHDI matrix's incompleteness, but they cannot precisely describe its symmetry.

Hence, above-mentioned models fail in either modeling its intrinsic symmetry or low-data density. Moreover, they consume expensive time cost to manually tune their multiple hyper-parameters for good performance. For addressing these crucial issues, this paper proposes a Proximal Symmetric Nonnegative Latent-factor-analysis (PSNL) model. The main contribution of this paper includes:

1. A PSNL model. It incorporates a proximal term into the symmetry-aware and data density-oriented objective function for high representation learning ability. For efficiently solving such an objective function, an adaptive Alternating Direction Method of Multipliers (ADMM)-based learning scheme is implemented through Tree-structured of Parzen Estimators (TPE) for high scalability.;

———————————————————————————————

- *Corresponding author: X. Luo.*
- *Y. Zhong, Z. Xie, W. Li and X. Luo are with the School of Computer Science and Technology, Dongguan University of Technology, Dongguan, Guangdong 523808, China (e-mail: zhongyurong91@gmail.com, gxyz4419@gmail.com, weilinglicq@outlook.com, luoxin21@gmail.com).*

2. It performs empirical studies on four real-world UWNs to demonstrate that a PSNL model outperforms state-of-the-art models in terms of representation learning ability, as well as highly competitive computational efficiency.

Section II introduces the preliminaries. Section III presents a PSNL model. Section IV gives the experimental results. Finally, Section V concludes this paper.

## II. PRELIMINARIES

### A. Problem Formulation

An SHDI matrix $Y$ generated by an UWN is as defined next [10, 11].

**Definition 1.** Given $U$, each entry quantifies some kind of interactions among the matrix $Y^{|U|\times|U|}$, which is non-negative. Given known set $\Lambda$ and unknown set $\Gamma$ for $Y$, $Y$ is an SHDI matrix if $|\Lambda|\ll|\Gamma|$.

For extracting potential yet useful information from an SHDI matrix, an NLF model is defined as:

**Definition 2.** Given $Y$ and $\Lambda$, an NLF model usually relies on $\Lambda$ to seek for rank-$f$ approximation of $Y$, i.e., $\hat{y}_{m,n}=a_{m,d}*b_{n,d}$ where $a_{m,d}\geq 0$ and $b_{n,d}\geq 0$. With commonly-used Euclidean distance [25-28, 38-40] and $L_2$-norm-based regularization scheme [21, 24, 36, 37], the following objective function is defined as:

$$F = \frac{1}{2}\sum_{y_{m,n}\in\Lambda}\left(\left(y_{m,n}-\sum_{d=1}^{f}a_{m,d}b_{n,d}\right)^2 + \lambda\sum_{d=1}^{f}\left((a_{m,d})^2+(b_{n,d})^2\right)\right), \quad (1)$$

$$s.t. \quad \forall m,n\in\{1,2,\ldots,|N|\}, d\in\{1,2,\ldots,f\}: a_{m,d}\geq 0, b_{n,d}\geq 0,$$

where the $L_2$-norm-based regularization coefficient $\lambda$ is positive.

## III. A PSNL MODEL

As shown in (1), an NLF model is not designed for $Y$'s symmetry. To address this issue, making $A=B$ in (1) is able to transform NLF to be a symmetry-aware model. Hence, the objective function $F$ can be reformulated as:

$$F = \frac{1}{2}\sum_{y_{m,n}\in\Lambda}\left(\left(y_{m,n}-\sum_{d=1}^{f}a_{m,d}a_{n,d}\right)^2 + \lambda\sum_{d=1}^{f}\left((a_{m,d})^2+(a_{n,d})^2\right)\right), \quad (2)$$

$$s.t. \quad \forall m,n\in\{1,2,\ldots,|U|\}, d\in\{1,2,\ldots,f\}: a_{m,d}\geq 0, a_{n,d}\geq 0.$$

Then, according to the previous study [49], a nonnegative constraint applied to the output LFs affects the model's representation accuracy. Hence, we introduce $X^{|U|\times f}$ into (2) to separate the nonnegative constraint from generalized loss:

$$F = \frac{1}{2}\sum_{y_{m,n}\in\Lambda}\left(\left(y_{m,n}-\sum_{d=1}^{f}x_{m,d}x_{n,d}\right)^2 + \lambda\sum_{d=1}^{f}\left((x_{m,d})^2+(x_{n,d})^2\right)\right),$$

$$s.t. \quad \forall m,n\in\{1,2,\ldots,|U|\}, d\in\{1,2,\ldots,f\}: a_{m,d}\geq 0, a_{n,d}\geq 0, \quad (3)$$

$$X = A.$$

According to the main principle of ADMM [49], Lagrangian multiplier matrix $W^{|U|\times f}$ is introduced for the equality constraint $X=A$, thereby achieving the following augmented Lagrangian function $\varepsilon$:

$$\varepsilon = \frac{1}{2}\sum_{y_{m,n}\in\Lambda}\left(\left(y_{m,n}-\sum_{d=1}^{f}x_{m,d}x_{n,d}\right)^2 + \lambda\sum_{d=1}^{f}\left((x_{m,d})^2+(x_{n,d})^2\right)\right)$$

$$+\sum_{u=1}^{|U|}\sum_{d=1}^{f}w_{u,d}\left(x_{u,d}-a_{u,d}\right) + \frac{\alpha_u}{2}\sum_{u=1}^{|U|}\sum_{d=1}^{f}\left(x_{u,d}-a_{u,d}\right)^2, \quad (4)$$

$$s.t. \quad \forall m,n\in\{1,2,\ldots,|U|\}, d\in\{1,2,\ldots,f\}: a_{m,d}\geq 0, a_{n,d}\geq 0,$$

where $\alpha_u$ controls the augmentation effects and it is set as $\alpha_u=\gamma|\Lambda(u)|$.

Afterwards, following previous studies [20, 51], proximal terms are able to improve representation learning ability of the model designed by the main principle of ADMM. Hence, we introduce a proximal term related to $X$ into (4), thereby reformulating (4) as follows:

$$\varepsilon = \varepsilon\left(X^k, A^k, W^k\right) + \frac{\mu}{2}\sum_{u=1}^{|U|}\sum_{d=1}^{f}\left(x_{u,d} - x_{u,d}^k\right), \quad (5)$$

$$s.t. \quad \forall m,n \in \{1,2,\ldots,|U|\}, d \in \{1,2,\ldots,f\}: a_{m,d} \geq 0, a_{n,d} \geq 0,$$

where $X^k$, $A^k$ and $W^k$ denote the statuses of $X$, $A$ and $W$ at the $k$-th iteration, and proximal coefficient $\mu$ is positive.

Following solution algorithms of an augmented Lagrangian function [19, 49], $\forall m, n \in \{1,2, \ldots, |U|\}, d \in \{1,2,\ldots,f\}$. Hence, the learning rules for $X$, $A$ and $W$ are given as:

$$x_{m,d}^{k+1} = \frac{\sum_{n \in \Lambda(m)}\left(y_{m,n} - \sum_{l=1, l \neq d}^{f} x_{m,l}^k x_{n,l}^k\right)x_{n,d}^k + \alpha_m a_{m,d}^k - w_{m,d}^k + \mu x_{m,d}^k}{\sum_{n \in \Lambda(m)}\left(\left(x_{n,d}^k\right)^2 + \lambda\right) + \alpha_m + \mu}, \quad (6a)$$

$$a_{m,d}^{k+1} = \max\left(0, x_{m,d}^{k+1} + w_{m,d}^k/\alpha_m\right), \quad (6b)$$

$$w_{m,d}^{k+1} = w_{m,d}^k + \eta\alpha_m\left(x_{m,d}^{k+1} - a_{m,d}^{k+1}\right), \quad (6c)$$

where the rescaling of learning rate $\eta$ is positive. Note that (6b) adopts nonnegative truncation method for ensuring $a_{m,d} \geq 0$.

Based on (6), $\forall d \in \{1,2,\ldots,f\}$, following a standard ADMM-incorporated learning sequence [49], PSNL splits the whole training task into $f$ disjoint subtasks, and then designs the following sequence for the $d$-the subtask whose composed of three jobs:

**1. Job One:**

$$X_{\cdot,d}^{k+1} \xleftarrow{(6a)} \arg\min_{X_{\cdot,d}} \varepsilon\left(\left[X_{\cdot,1\sim(d-1)}^{k+1}, X_{\cdot,d}, X_{\cdot,(d+1)\sim f}^{k}\right], A_{\cdot,d}^k, W_{\cdot,d}^k\right), \quad (7a)$$

**2. Job Two:**

$$A_{\cdot,d}^{k+1} \xleftarrow{(6b)} \arg\min_{A_{\cdot,d}} \varepsilon\left(X_{\cdot,d}^{k+1}, A_{\cdot,d}, W_{\cdot,d}^k\right), \quad (7b)$$

**3. Job Three:**

$$W_{\cdot,d}^{k+1} \xleftarrow{(6c)} W_{\cdot,d}^k + \eta\alpha_m \nabla_{W_{\cdot,d}} \varepsilon\left(X_{\cdot,d}^{k+1}, A_{\cdot,d}^{k+1}, W_{\cdot,d}^k\right). \quad (7c)$$

Note that it is crucial to find optimal the model's hyper-parameters for achieving its good performance [14, 32, 41]. According to previous studies [50], we adopt one kind of BO method, i.e., the commonly-adopted tree-structured of parzen estimators (TPE) algorithm for implementing adaptation of PSNL's hyper-parameters, i.e., $\lambda$, $\gamma$, $\mu$ and $\eta$. Let $s=\{\lambda, \gamma, \mu, \eta\}$ and $b$ denotes the loss computed by $y_{m,n}$ and $\hat{y}_{m,n}$ with $s$. Given the observation set $C=\{(s_1, b_1), (s_2, b_2), ..., (s_\beta, b_\beta)\}$, TPE seeks to optimize the following Expected Improvement (EI):

$$EI_{b^*}(s) \propto \left(\theta + g(s)/l(s)*(1-\theta)\right). \quad (8)$$

According to (8), TPE returns $s$ related to the minimal $b$ value in $C$. Note that TPE is implemented by "Hyperopt" software package in [50].

## IV. EXPERIMENTAL RESULTS AND ANALYSIS

*A. General Settings*

**Evaluation Protocol.** In real-world applications, decomposing an SHDI matrix into LFs to predict missing values is critical since discovering connections between potentially involved entities is highly desired [10, 11, 22]. As a result, we employ it as an evaluation protocol to test the performance of related models.

**Evaluation Metrics.** The accuracy of a test model for missing data predictions can be measured by the root mean square error (RMSE) [15-18, 34, 35, 52]:

$$RMSE = \sqrt{\left(\sum_{r_{i,j} \in \Gamma}(r_{i,j} - \hat{r}_{i,j})^2\right)/|\Gamma|},$$

where $\Gamma$ denotes the validation set and is disjoint with the training set $\Lambda$. Note that low RMSE represents high prediction accuracy for missing data in $\Gamma$. Meanwhile, time cost of each involved model is recorded for testing its computational efficiency.

TABLE I. Details of Adopted Datasets.

| No. | Type | $|\Lambda|$ | $|N|$ | Density | Source |
|---|---|---|---|---|---|
| D1 | Protein | 1,120,028 | 7,963 | 1.77% | [46] |
| D2 | Protein | 1,021,786 | 4,181 | 5.85% | [46] |
| D3 | Material | 322,905 | 13,965 | 0.17% | [47] |
| D4 | Knowledge | 57,002 | 2,427 | 0.97% | [47] |

TABLE II. Details of Tested Models.

| No. | Name | Description |
|---|---|---|
| M1 | NMFC | An NMF model solved by a standard ADMM [49]. |
| M2 | NIR | A recent improved NLF model with regularization effects connected with data density [48]. |
| M3 | $\beta$-SNMF | An SNMF model solved by an improved NMU-based algorithm [4]. |
| M4 | GSNMF | An SNMF model with graph regularization and solved by an improved NMU-based algorithm [5]. |
| M5 | LightGCN | A simplified GCN model [2]. |
| M6 | PSNL | The proposed model in this paper. |

TABLE III. RMSE of M1-6 on D1-4.

| No. | M1 | M2 | M3 | M4 | M5 | M6 |
|---|---|---|---|---|---|---|
| D1 | $0.1329_{\pm 3.8E\text{-}4}$ | $0.1274_{\pm 7.9E\text{-}4}$ | $0.1652_{\pm 6.3E\text{-}4}$ | $0.1813_{\pm 1.8E\text{-}4}$ | $0.1297_{\pm 1.4E\text{-}4}$ | $\mathbf{0.1266}_{\pm 9.1E\text{-}5}$ |
| D2 | $0.1334_{\pm 4.1E\text{-}4}$ | $0.1291_{\pm 4.2E\text{-}4}$ | $0.1543_{\pm 2.8E\text{-}4}$ | $0.1576_{\pm 1.5E\text{-}4}$ | $0.1318_{\pm 1.6E\text{-}4}$ | $\mathbf{0.1278}_{\pm 6.0E\text{-}5}$ |
| D3 | $0.0821_{\pm 2.1E\text{-}5}$ | $0.0751_{\pm 3.8E\text{-}4}$ | $0.0741_{\pm 1.1E\text{-}5}$ | $0.0764_{\pm 8.4E\text{-}5}$ | $0.1037_{\pm 2.2E\text{-}4}$ | $\mathbf{0.0736}_{\pm 1.5E\text{-}4}$ |
| D4 | $0.0971_{\pm 1.3E\text{-}3}$ | $0.0705_{\pm 5.2E\text{-}4}$ | $0.1093_{\pm 6.5E\text{-}5}$ | $0.1255_{\pm 1.5E\text{-}3}$ | $0.0928_{\pm 1.1E\text{-}3}$ | $\mathbf{0.0631}_{\pm 2.8E\text{-}4}$ |

TABLE IV. Time Cost of M1-6 on D1-4 (Seconds).

| No. | Type | M1 | M2 | M3 | M4 | M5 | M6 |
|---|---|---|---|---|---|---|---|
| D1 | Tuning | $136{,}172_{\pm 10{,}672.45}$ | $8{,}461_{\pm 796.09}$ | $7{,}810_{\pm 793.26}$ | $52{,}077_{\pm 4{,}338.75}$ | $77{,}453_{\pm 5902.87}$ | $\mathbf{3{,}705}_{\pm 309.18}$ |
| D1 | Testing | $2{,}477_{\pm 358.66}$ | $\mathbf{33}_{\pm 9.61}$ | $4{,}947_{\pm 471.63}$ | $5{,}447_{\pm 568.44}$ | $562_{\pm 41.72}$ | $48_{\pm 41.72}$ |
| D2 | Tuning | $24{,}550_{\pm 1963.27}$ | $5{,}349_{\pm 473.24}$ | $\mathbf{1{,}569}_{\pm 135.89}$ | $8{,}532_{\pm 916.53}$ | $60{,}628_{\pm 7068.25}$ | $2{,}325_{\pm 186.73}$ |
| D2 | Testing | $336_{\pm 29.12}$ | $\mathbf{23}_{\pm 6.78}$ | $995_{\pm 72.51}$ | $2{,}176_{\pm 627.61}$ | $551_{\pm 36.03}$ | $64_{\pm 7.96}$ |
| D3 | Tuning | $398{,}144_{\pm 23{,}807.64}$ | $1{,}455_{\pm 135.12}$ | $21{,}606_{\pm 2002.83}$ | $148{,}565_{\pm 12{,}473.96}$ | $8{,}342_{\pm 709.83}$ | $\mathbf{821}_{\pm 71.46}$ |
| D3 | Testing | $5{,}032_{\pm 488.67}$ | $105_{\pm 26.39}$ | $786_{\pm 64.77}$ | $2{,}781_{\pm 578.66}$ | $\mathbf{51}_{\pm 7.91}$ | $66_{\pm 5.12}$ |
| D4 | Tuning | $14{,}284_{\pm 1142.95}$ | $415_{\pm 58.91}$ | $1{,}130_{\pm 130.59}$ | $6{,}324_{\pm 546.24}$ | $2{,}112_{\pm 161.72}$ | $\mathbf{412}_{\pm 38.03}$ |
| D4 | Testing | $521_{\pm 48.21}$ | $6_{\pm 0.71}$ | $65_{\pm 7.17}$ | $278_{\pm 57.12}$ | $83_{\pm 10.79}$ | $\mathbf{2}_{\pm 0.08}$ |

Tuning denotes tuning time cost; Testing denotes testing time cost.

**Datasets.** Our experiments adopt four UWNs, and their details are shown in Table I. In all experiments on each dataset, we randomly split its known set of entries Λ into ten disjoint subsets for tenfold cross-validation. More specifically, each time we adopt seven subsets as the training set, one subset as the validation set, and the remaining two subsets as the test set. This process is repeated ten times sequentially.

**Compared Models.** Our experiments involve six models, and their details are presented in Table II. To achieve its objective results, we used the following settings:

(1) LF Dimension $d$ is set at 20;
(2) For each model on each data set, the results generated from 10 different random initial values are recorded to calculate the average RMSE and convergence time for eliminating the effect of initial assumptions [13, 42-45].
(3) The training process of the test model is terminated when: 1) the iteration count reaches a preset threshold, which is 1000; 2) The difference between the two consecutive iterations of the generated RMSE is less than $10^{-5}$.

*B. Comparison against State-of-the-art Models*

Table III and IV summarize RMSE and time cost of M1-6 on D1-4, respectively. From these results, we have the following findings:

(1) **PSNL achieves significantly higher accuracy gain than state-of-the-art models do.** For example, as shown in Table III, RMSE of M6 is 0.0631 on D4, which is about 35.02%, 10.5%, 42.27%, 49.72% and 32% lower than M1's 0.0971, M2's 0.0705, M3's 0.1093, M4's 0.1255 and M5's 0.0928, respectively. Similar results can be found on D1-3. Hence, PSNL achieves significantly higher representation accuracy than state-of-the-art models do.
(2) **PSNL's computational efficiency is higher than state-of-the-art models do.** Owing to data density-oriented modeling and adaptive learning scheme, PSNL's computational efficiency is highest, i.e., its total time cost (Tuning + Testing) is lowest, as shown in Table IV.

## V. Conclusions

A PSNL model has shown great potential in predicting missing data of an SHDI matrix generated by an UWN. Its high representation accuracy and hyper-parameter adaptation highly boost its practicability. The research of representation learning of an UWN remains in its infancy, and its applications have potential development, which is worthy of further investigation in the future.